\documentclass[conference]{IEEEtran}

\usepackage{cite}
\usepackage{amsmath,amssymb,amsfonts,amsthm}
\usepackage{algorithmic}
\usepackage{graphicx}
\usepackage{textcomp}
\usepackage{xcolor}
\newtheorem{definition}{Definition}

\RequirePackage[utf8]{inputenx}
\def\BibTeX{{\rm B\kern-.05em{\sc i\kern-.025em b}\kern-.08em
    T\kern-.1667em\lower.7ex\hbox{E}\kern-.125emX}}

\usepackage{booktabs} 
\usepackage{bm}
\usepackage{dsfont}
    
\begin{document}

\title{Robust commuter movement inference from connected mobile devices}

\author{\IEEEauthorblockN{  Baoyang Song }
\IEEEauthorblockA{\textit{Ecole Polytechnique } \\
Palaiseau, France \\
baoyang.song@polytechnique.edu }
\and
\IEEEauthorblockN{  Hasan Poonawala }
\IEEEauthorblockA{\textit{ IBM Research  } \\
Singapore \\
hasanp@sg.ibm.com  }
\and
\IEEEauthorblockN{  Laura Wynter }
\IEEEauthorblockA{\textit{ IBM Research  } \\
Singapore \\
lwynter@sg.ibm.com  }
\and
\IEEEauthorblockN{  Sebastien Blandin }
\IEEEauthorblockA{\textit{ IBM Research  } \\
Singapore \\
sblandin@sg.ibm.com  }
}
\maketitle

\begin{abstract}
The preponderance of connected devices provides unprecedented opportunities for fine-grained monitoring of the public infrastructure. However while classical models expect high quality application-specific data streams, the promise of the \textit{Internet of Things} (IoT) is that of an abundance of disparate and noisy datasets from connected devices. In this context, we consider the problem of estimation of the level of service of a city-wide public transport network. We first propose a robust unsupervised model for train movement inference from wifi traces, via the application of robust clustering methods to a one dimensional spatio-temporal setting. We then explore the extent to which the demand-supply gap can be estimated from connected devices. We propose a classification model of real-time commuter patterns, including both a batch training phase and an online learning component. We describe our deployment architecture and assess our system accuracy on a large-scale anonymized dataset comprising more than $10$ billion records.
\end{abstract}

\begin{section}{Introduction}\label{sec:intro}
Artificial Intelligence methods have seen tremendous progress in  recent years, as illustrated by the significant progress made on curated and customized datasets and in the context of games. In practice, however, data streams are rarely curated in advance and exhibit a number of properties making it inadequate for off-the-shelf machine learning methods.

In this work, we consider the problem of fine-grained monitoring of public transport systems, via the creation of a \textit{digital twin} of the system, able to faithfully capture its fundamental properties. While traditional public infrastructure monitoring has historically been based on proprietary and application-specific data streams, the democratization of connected devices is a considerable game changer, making available datasets that are orders of magnitude larger and noisier.

We focus our efforts on assessing the extent to which key performance indicators and real-time operational information needed for monitoring public transport systems can be obtained with sufficient accuracy from passive sensing via the use of appropriate machine learning techniques. One example is determining the true (as opposed to the published) train arrival timings of a public transport system in real-time. Often, there is no publicly available source for this information, yet, a reliable timetable is indispensable for the real-time monitoring of public transport service levels, and is of great value to commuters when planning their trips.

Another important metric to assess service quality of public transport is the \textit{demand-supply gap} (DSG), evidenced by the inability for a passenger to board a train due to the train being already at capacity when it enters a station. These events signal an inadequacy of the supply with respect to the demand and as such are an effective way to measure public transport service quality. Historically, demand-supply gap has been evaluated from necessarily sub-sampled surveys conducted manually and available offline compared to the events they refer to.
\begin{subsection}{Literature review}
There has been considerable work in recent years on understanding mobility patterns using data from mobile devices~\cite{ma2014opportunities, jiang2016timegeo, poonawala2016singapore,caspari2017real,shao2016slicing,calabrese2015urban, jiang2017activity, candia2008uncovering}, and using them in the context of incident management~\cite{szabo2017data}. These works often make use of telecommunications data to trace movements of people. However, spatial resolution of the telecommunications data is usually quite coarse. GPS signal enables fine-grained estimation~\cite{herrera2010evaluation}. In the context of public transport, wifi traces have the advantage of having very fine spatial resolution, and do not require a proprietary sensing mechanism.

In~\cite{paper-51}, the authors design a system to estimate the number of passengers in public transport vehicles. The authors of~\cite{6583443} build a system on top of a Raspberry Pi to track users' locations at a mass event using probe, association and re-association requests. In~\cite{baoyang}, the authors build a system to ``sniff'' wifi signals of office workers along with an online SVM-based model to predict their lengths of stay.

In~\cite{Wang:2014:THQ:2594368.2594382}, the authors study queue waiting times using single-point wifi and propose a Bayesian network to estimate queue length. The authors of~\cite{6567123} design a dwell time prediction framework for  retail store environments using various sensors from smartphones including wifi signal strength and data transmission rate;  their approach relies, however, on features requiring  users to install dedicated software on their devices, limiting the applicability of the framework. 

In an offline setting,~\cite{Rose:2010:MUW:1869983.1870015} leverage probe requests to reveal past behaviors of users. Similarly,~\cite{6415713} employ the spatio temporal information of probe requests to reveal the underlying social relationships within a small sample of users. In~\cite{Barbera:2013:SCU:2504730.2504742} the authors build snapshots of users involved in a large scale event.

The problem of inferring or predicting train arrivals and hence delays of public transport services has been addressed in a number of studies, assuming access to train locations from, for example, the train signaling system. In practice, though, only the train operator has access to the actual train location data. This gap is partially addressed by~\cite{HORN201567} which derives regional train timetables using cell phone data by detecting bursts in number of cell phone subscribers. The authors report a precision of $85\%$ within $5$ minutes, but a recall rate of only $49\%$. While their method works reasonably well for highly separated regional train lines, it would not work well on a dense urban metro system. In addition, as with train signaling data, cell phone records are not readily available. 

The problem of detecting events of commuters \textit{left behind} in a subway system is addressed in~\cite{isttt}. The authors rely on offline farecard data, and estimate a model using a maximum likelihood approach assuming known distributions of waiting times and walking times of the passengers. 
\end{subsection}
\begin{subsection}{Contributions}
Our contribution is thus to define means for using passive, universally-accessible wifi data to infer train movements and train service levels. More specifically:
\begin{itemize}
	\item we formulate the problem of fine-grained passive sensing and illustrate the associated challenges,
	\item we propose unsupervised and supervised models for train movement inference and demand-supply gap estimation that are robust to the inherent limitations of wifi-based sensing,
	\item we deploy our models within a Big Data architecture handling city-scale data volume on the order of hundreds of stations and millions of users in real-time,
	\item we conduct a thorough model evaluation and present accuracy results.
\end{itemize}
\end{subsection}

The rest of the article is organized as follows: Section~\ref{sec:prob} formulates the problem and goes over the challenges associated with sensing from connected devices. Sections~\ref{sec:mrt:stationwise} and~\ref{sec:models} present the spectral clustering approach for train movement inference, and the demand-supply gap learning model, respectively.  Section~\ref{sec:mrt:evaluation} describes our numerical results and concluding remarks are provided in Section~\ref{sec:conc}.
\end{section}
\begin{section}{Problem statement}\label{sec:prob}
In this section we outline the specificities of the problem of real-time estimation from connected devices.
\begin{subsection}{Passive sensing}
The ubiquity of connected devices creates the conditions for large-scale decentralized open monitoring systems. Wifi networks are probably exemplary of this paradigm, since wifi access points (AP) are omnipresent and most mobile devices support the wifi protocol. 

In order to discover and automatically connect to known wifi networks, mobile stations scan periodically the wifi bands by broadcasting on all available channels their so-called \emph{probe requests}~\cite{cunche:hal-00874078,paper-51}, providing information of the spatio-temporal locations of connected devices. It is important to note that even if no AP is present, probe requests are systematically sent and thus connected devices are observable.

In fact, thanks to the openness of the protocol, it is possible without proprietary technical knowledge, to collect these probe requests by using a wifi sniffer such as \texttt{tcpdump} or  \texttt{Wireshark}. In addition, accessing probe requests is device-agnostic and non-intrusive. 

On the other hand, because the wifi protocol stems from the need for efficient wireless communication, it does not  exhibit properties conducive to accurate and effective sensing. For instance it is clear that probe requests are correlated with the device activity, and hence inherently \textit{event-based}, yet the event is of no immediate relevance to spatio-temporal sensing. Additionally, for a variety of reasons, device identifiers are sometimes randomized, leading to further noise in the traces.

In the following section, we go over the main challenges associated with passive sensing for fine-grained public infrastructure monitoring.
\end{subsection}
\begin{subsection}{Wifi-based public transport monitoring}
In this work, we assume that wifi access points are located at the platforms of a train network so that wifi connectivity occurs when a device is physically on or near a railway platform. A record consists of a device identifier, a location expressed in a specific system of coordinates, and a timestamp.

Because devices scan for access points at a frequency (typically $10-100$ Hz) much greater than the frequency of relevant public transport events (typically $0.1-0.01$ Hz), we first pre-process the raw data and identify a journey with: 
\begin{equation}
 \bm{t} = ((t^{A}_{1},t^{D}_{1}),\ldots,(t^{A}_{N_{i}},t^{D}_{N_{i}}))_{i\in \mathbb{N}}
 \label{eqn:mrt:vectorization}
\end{equation}
where $i$ denotes the device id, and $(t^{A}_{j},t^{D}_{j})$ denote the first and last times when the device is observed at location $j$. By ignoring the events that are not the first or last event at a given location, we significantly reduce the amount of data to be processed (by several orders of magnitude) and preserve the spatio-temporal statistics of events of interest since we are considering location-aggregate quantities (where a location is typically a station). In this work we do not consider within-location spatially heterogeneous phenomena.
\begin{subsubsection}{Uncertain observation probability}
We first note that since the sensing mechanism is passive, there is no guarantee for a commuter to be observed. This can be due to a number of causes; variable frequency of probe requests, randomization of MAC addresses, lack of reliability of sensing or positioning scheme, etc. We illustrate this property in Figure~\ref{fig:mrt:example_journey}.
\begin{figure}[!htb]
 \centering
 \includegraphics[width = 0.8\columnwidth]{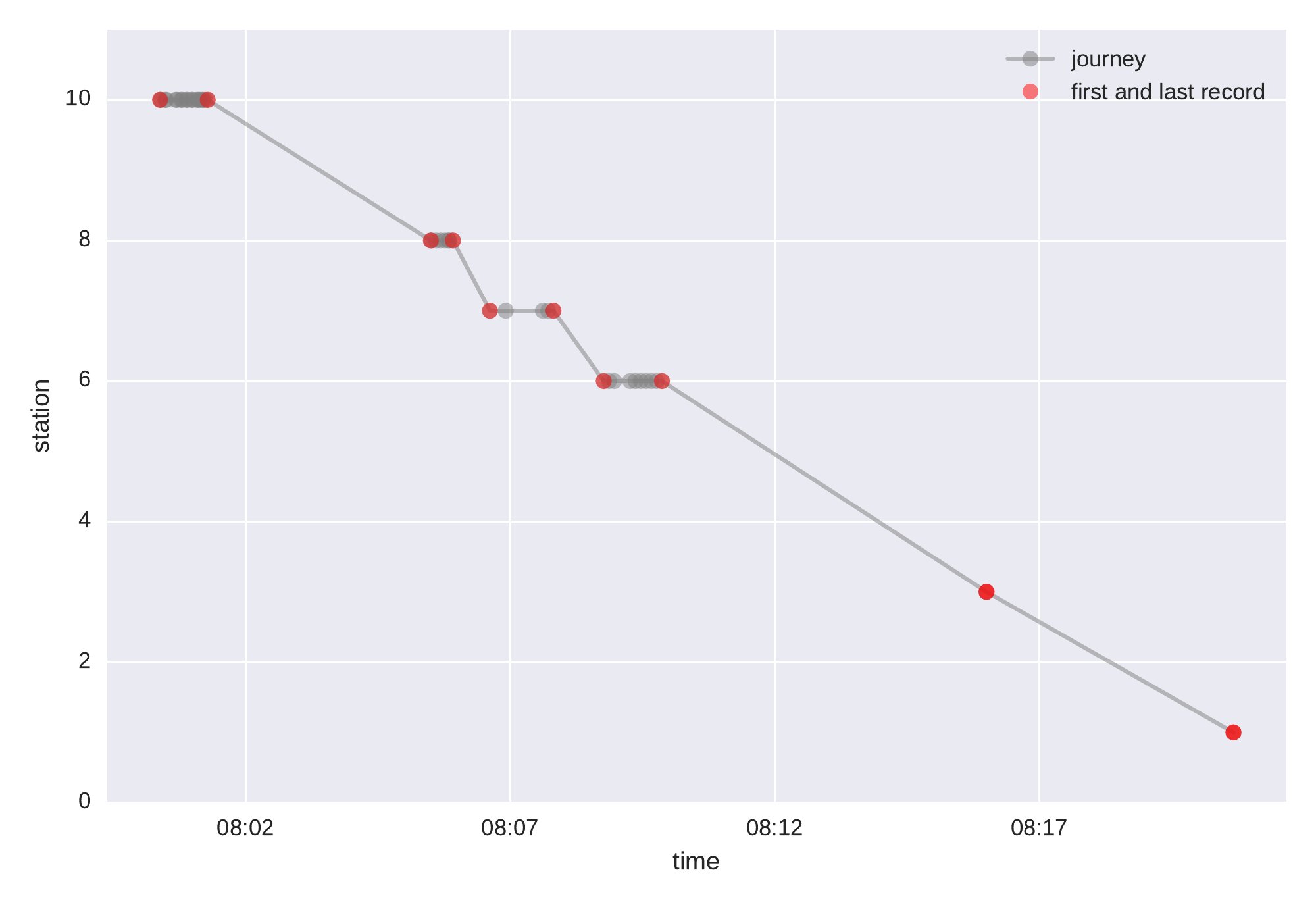}
 \caption{Journey of a connected device: through $10$ train stations. The first and the last wifi record observed at each station are colored in red.}
 \label{fig:mrt:example_journey}
\end{figure}

While it is clear using a continuity argument that the commuter associated with this device travels from station $10$ to station $1$, a significant proportion of observations are missing.
\end{subsubsection}
\begin{subsubsection}{Non-stationary sensing properties}
The second challenge associated with wifi-based sensing is the lack of consistency of the sensing scheme parameters. Considering for instance the proportion of commuters sampled by the wifi-based sensing scheme, one would expect a relatively stable proportion, only depending on the penetration of connected devices in the population considered.

We illustrate in Figure~\ref{fig:sampling} that the proportion of commuters observed is much less regular than one might expect and is both highly time-varying and location-dependent.
\begin{figure}[!htb]
 \centering
 \includegraphics[width=0.8\columnwidth]{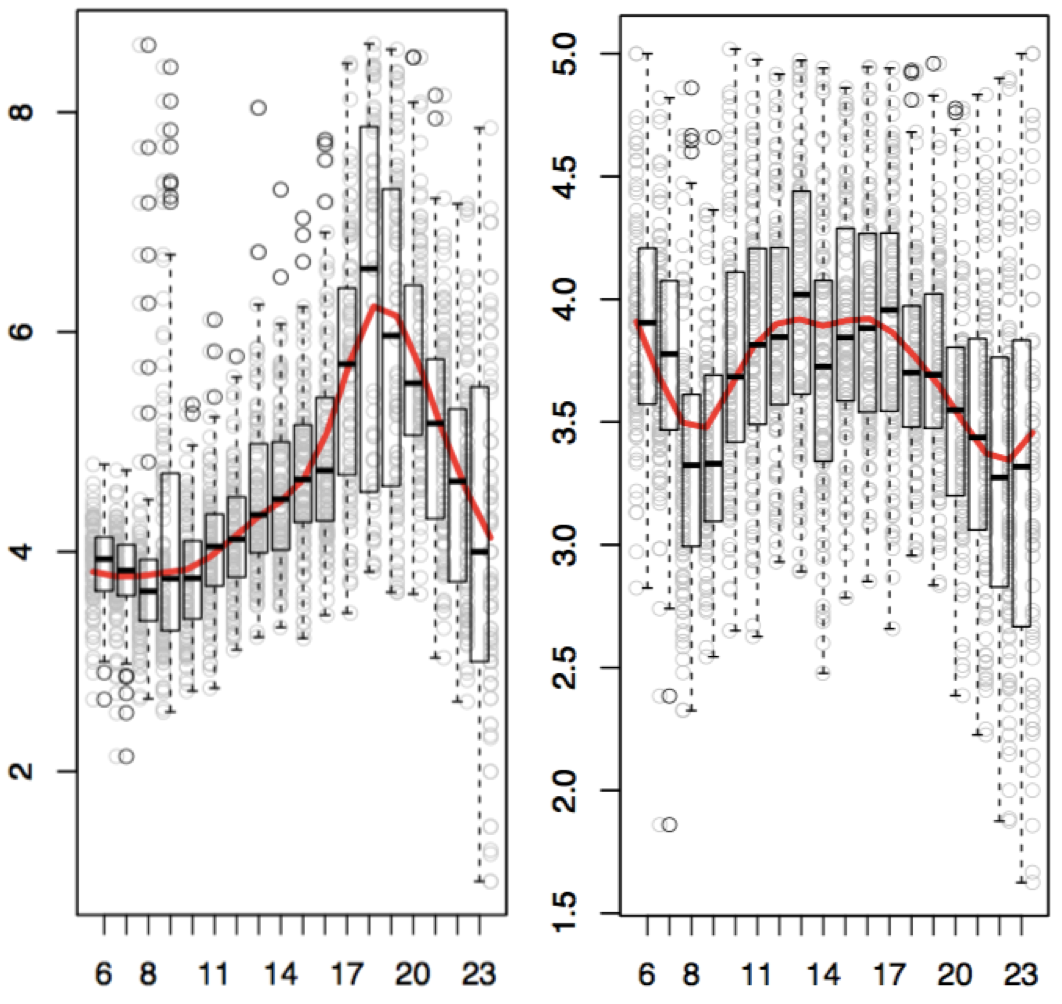}
 \caption{Observed proportion, i.e., number of commuters entering a given station divided by the number of connected devices observed at the same station. The x-axis is the hour of the day from 6am to 12am. The y-axis gives the proportion distribution over 4 weeks.  The station on the right shows a stable proportion while the station on the left has a high variance.}
 \label{fig:sampling}
\end{figure}
\end{subsubsection}
\begin{subsubsection}{Spatio-temporal accuracy limitations}
Finally we discuss the lack of accuracy of position estimates generated from connected devices traces. Figure~\ref{fig:mrt:mis-identified} presents a set of wifi traces across a set of $3$ stations over a $10$ minutes interval.
\begin{figure}[!htb]
 \centering
 \includegraphics[width=0.9\columnwidth]{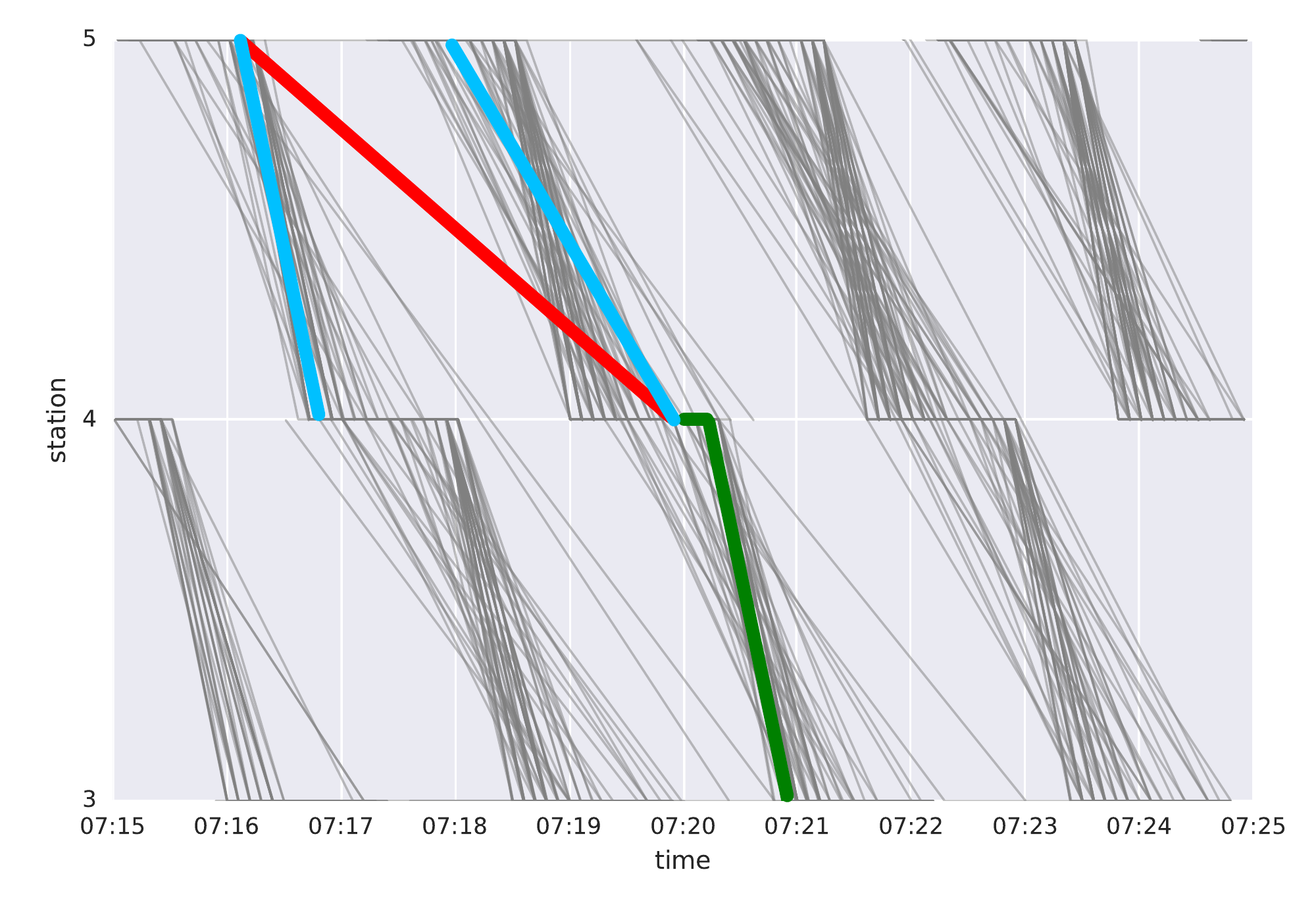}
 \caption{Location time uncertainty: illustrated by the red trace, corresponding to a non-physical trace boarding a train at station $5$ around $7:16$ but alighting from the preceding train at station $4$ around $7:20$.}
 \label{fig:mrt:mis-identified}
\end{figure}
While the blue traces are indicative of reliable traces because they qualitatively seem to correspond to distinct trace clusters suggesting distinct trains, it is clear that the red trace is non-physical, in that it reflects the movement of a device starting its trip at station $5$ around $7:16$, and then catching up with the previous train at station $4$ around $7:20$. This may be caused by the device not being sampled towards the end of its stay at station $5$. Such  examples illustrate the lack of accuracy in the spatial and temporal quantities provided by the individual records, and the need for robustness in the estimation algorithms considered.
\end{subsubsection}
\end{subsection}
\begin{subsection}{Big Data platform}
Our application is deployed as a real-time analytics platform in a city-scale context, ingests data from on the order of hundreds of stations, processes the data and runs the machine learning components described subsequently in real-time. Given operational targets, it is important for the real-time pipeline to scale, in particular during peak hours when the traffic is an order of magnitude greater than off-peak traffic.

We illustrate in Figure~\ref{fig:wifiMLPipeline} our real-time machine learning pipeline, architected around the IBM InfoSphere Streams processing engine able to seamlessly spawn new processing streams as required. The IBM Integration Bus handles mini-batch data records received from the on the ground sensing system, and we also use an in-memory database in order to minimize end-to-end latency.
\begin{figure}[!htb]
	\centering
	\includegraphics[width=0.9\columnwidth]{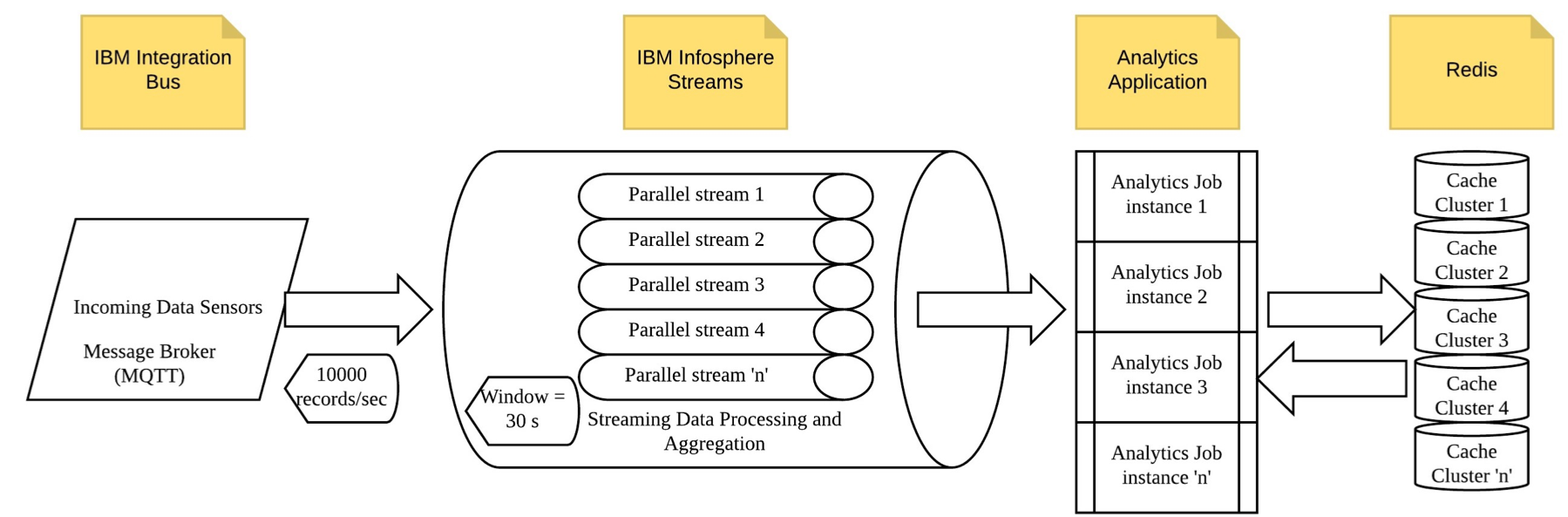}
	\caption{Real-time data analytics pipeline.}
	\label{fig:wifiMLPipeline}
\end{figure} 
Model training, not depicted here, follows a classical Spark architecture, in its IBM Big Insights implementation. We typically train the models using  $1$ year of wifi traces, i.e. more than $10$ billion records.
\end{subsection}
In the following section we present the machine learning models.
\end{section}
\begin{section}{Spectral clustering of noisy traces}\label{sec:mrt:stationwise}
In this section we introduce the unsupervised learning model used for train movement identification. We first present a baseline model and its limitations, motivating the use of a more sophisticated model.
\begin{subsection}{Baseline model}
The baseline method considers each station independently, performs station-wise clustering of event timestamps, and then re-identifies clusters across stations. For univariate time-series data, a well-known clustering method is DBSCAN~\cite{473950}. For robustness the final step consists of pruning the events not associated with train movements (i.e. for instance corresponding to idle commuters). The procedure is iterative: suppose that clusters at all stations $j$ such that $j < i$ are successfully identified. Given a station $i$ and a cluster $l_i$, if a record is seen previously in a cluster $l_j$ at a station $j (j < i)$, and the timestamp difference is within a tolerable range (in our experiments within $30$ minutes), we consider that $l_i$ should actually be labeled $l_j$. We examine each record and use majority voting to determine the final cluster label.

A limitation of the baseline method is that it can fail to distinguish distinct trains, especially at peak times when the headway is short. Consider the following example: a line with three stations $i, j$ and $k$ and a single train traveling from station $i$ to station $k$ via station $j$. Suppose that at station $j$ no passenger having boarded at $i$ is observed, but at station $k$, all passengers boarding at $i$ and $j$ are recognized. Given that none of the passengers observed at station $i$ is observed at station $j$, the baseline model considers the train going through station $j$ to be distinct from the train defined by commuters seen at both station $i$ and station $k$. This is an example of an erroneous detection, when a train is wrongly detected as distinct from the existing train set.

The opposite type of error is illustrated in Figure~\ref{fig:mrt:mis-identified}, and is associated with the fact that two traces are considered as belonging to the same cluster when they do not in practice. 
\end{subsection}
\begin{subsection}{Spatio-temporal embedding}
In order to address the limitations of the baseline model presented in the previous section, we propose an approach defined by a global view of reconstructed train trips. We first define a line-level similarity metric for any two journeys.
\begin{definition}[Soft and hard embeddings]
 \label{def:mrt:similarity}
 For $\bm{t}_1,\bm{t}_2$ two points as in (\ref{eqn:mrt:vectorization}), we define 
 \begin{align}
 \mathrm{sim}_\mathrm{soft}(\bm{t}_1, \bm{t}_2) & = \|\bm{t}_1 - \bm{t}_2\|_0
 \exp\left(-\frac{\|\bm{t}_1 - \bm{t}_2\|_{\infty}^2}{2\sigma^2}\right)\nonumber\\
 \mathrm{sim}_\mathrm{hard}(\bm{t}_1, \bm{t}_2) & = \|\bm{t}_1 - \bm{t}_2\|_0
 \mathds{1}_{\|\bm{t}_1 - \bm{t}_2\|_{\infty} \leq \tau}.
 \label{eqn:mrt:softAndHard}
 \end{align}
\end{definition}
 In~\eqref{eqn:mrt:softAndHard}, the $L^{0}$ term quantifies the spatial similarity, \emph{i.e.}\ the number of stations where both journeys are recorded, the infinite norm term $L^{\infty}$ quantifies the temporal similarity, \emph{i.e.}\ the maximum time difference at stations where both  journeys are recorded. Note that Definition~\ref{def:mrt:similarity} is robust to \emph{mis-identified} journeys, that is, records belonging to two different trains but identified as a single journey. Figure~\ref{fig:mrt:mis-identified} shows such an example (the journey in red). According to Definition~\ref{def:mrt:similarity}, the blue journeys are dissimilar to the red journey due to the infinite norm term in~\eqref{eqn:mrt:softAndHard}. The red journey thus has very low similarity with the blue journeys, and the two trains to which the two blue journeys belong are unlikely to be wrongly assigned to a single cluster due the red journey.
\end{subsection}
\begin{subsection}{Robust spectral clustering}
Spectral clustering considers the entire journey of a traveler, hence in the context of our application, can contribute to addressing issues stemming from the local nature of the baseline model. For more details on  spectral clustering, we refer the interested reader to the  tutorial of~\cite{vonLuxburg2007}.

One of the difficulties with spectral clustering lies in choosing \emph{a-priori} the number of clusters $k$. However, the similarity graph associated with the embedding from Definition~\ref{def:mrt:similarity} is usually very sparse, therefore, we use the \emph{eigengap} heuristic~\cite{vonLuxburg2007}, which consists of finding the largest gap between the eigenvalues of the Laplacian matrix, to determine \emph{a-priori} the optimal value of $k$. Figure~\ref{fig:mrt:spectral-soft} shows the result of spectral clustering with soft similarity metric for our toy problem, and using the \emph{eigengap} heuristic to define the number of clusters.
\begin{figure}[!htb]
    \centering
    \includegraphics[width =0.9\columnwidth]{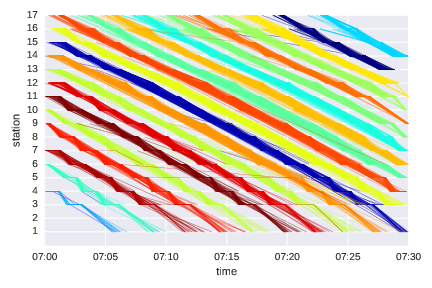}
    \caption{Spectral clustering applied to toy problem. Each color represents a cluster label. Soft similarity, $2\sigma^2 = 30$.}
  \label{fig:mrt:spectral-soft}
\end{figure}

We observe that most of the model errors are associated with distinct clusters being wrongly associated. The converse issue of clusters being fragmented is much less common because from the spectral clustering output and centrality estimates on travel-times, cluster fragments are easily reconciled, hence do not require complex special treatment. In order to address the issue of clusters being wrongly associated, for example the cluster at 7:00 at station $13$ and the cluster at 7:00 at station $10$ being wrongly connected, we post-process the clustering results to eliminate wrong cluster associations, and specifically the following two types of outliers:
\begin{itemize}
\item data points whose label is different from their neighbors label : we remove these  by thresholding labels of a k-NN model. This approach is not only robust to mis-classified journeys, but also to having too many clusters k.
\item data points who are not similar to their neighbors: we remove these by thresholding the similarity metric for each cluster.
\end{itemize}
The resulting clusters with outlier detection are shown in Figure~\ref{fig:mrt:denoised-envelope}.
\begin{figure}[!htb]
 \centering
 \includegraphics[width =0.9\columnwidth]{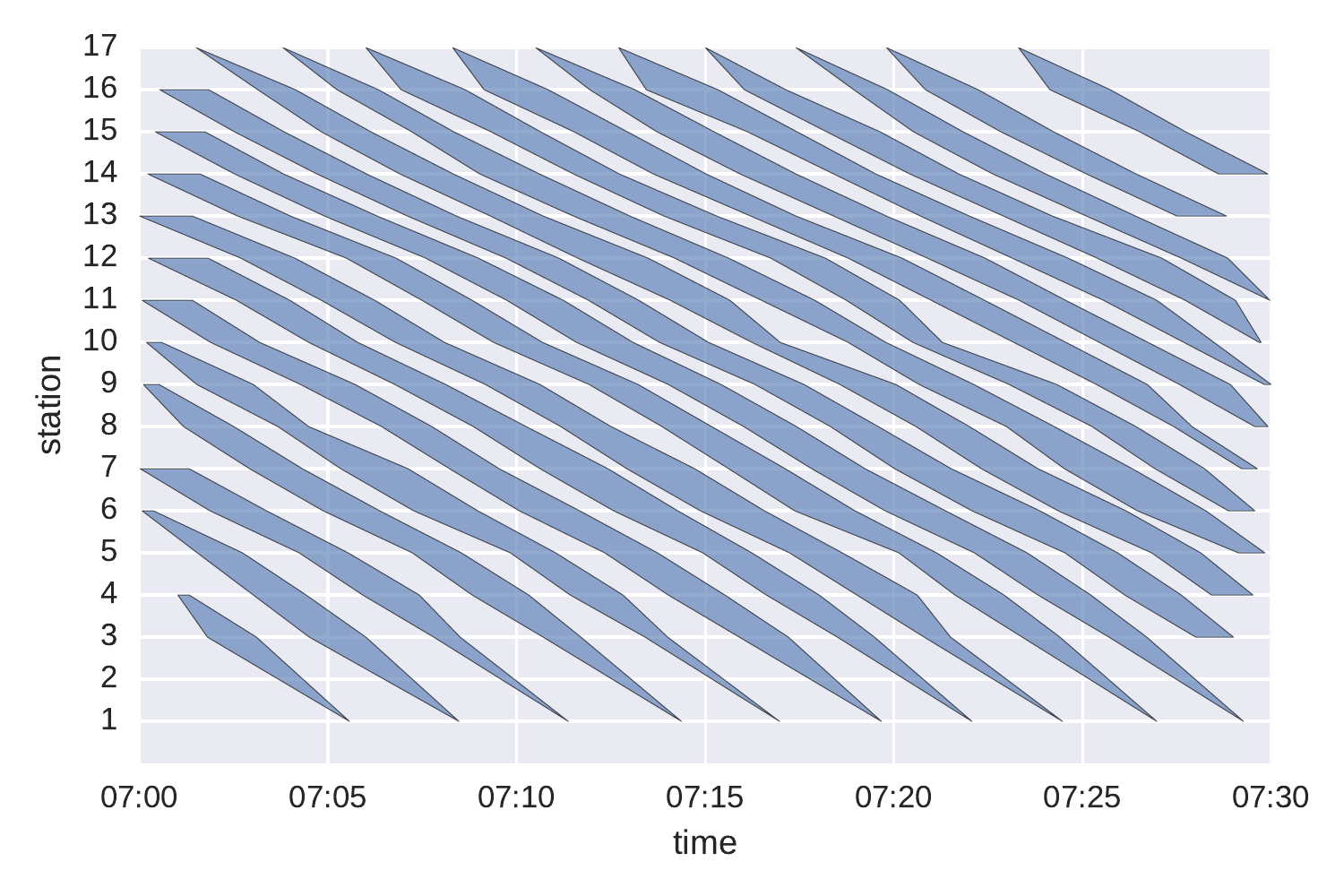}
 \caption{Envelope of train arrivals and departures based on spectral clustering and outlier detection algorithms.}
 \label{fig:mrt:denoised-envelope}
\end{figure}
As we can see, most wrongly associated clusters are detected and removed.
\end{subsection}
In the following section we present a model using the spectral clustering output in order to provide estimates of the demand-supply gap.
\end{section}
\begin{section}{Inference model for demand-supply gap estimation}\label{sec:models}
In this section we first present the intermediary processes used to estimate relevant features, and then describe the core model for demand-supply gap estimation.
\begin{subsection}{Preliminaries}
The count of passengers waiting to board a train at any given point in time is an informative albeit indirect measure of the unmet demand. However no sensor provides a complete measurement of that quantity. In particular CCTV  provides observations on portions of the platform and is notoriously difficult to use for measuring accurately the demand-supply gap. Ticketing data  provides only the entry counts,   reflecting the demand rather than the demand-supply gap.

In order to estimate the count of passengers waiting to board a train, given that noisy observations of the number of connected devices are available, we propose to learn   scaling factors relating the number of connected devices observations to the number of commuters. As illustrated in Figure~\ref{fig:sampling}, this number is both time-varying and station-dependent.

Assuming that the scaling factor is uniform within each station considered. Let  $Y$ be the count of passengers entering the station as observed by the fare gate sensors, with $X$ the count of passengers entering the station as observed from connected devices, for a station $s$, and time $t$, the scaling factor $\theta_{s,t}$ reads:
\begin{equation}\label{scale_factor}
\theta_{s,t} = \frac{Y_{s,t}}{X_{s,t}}.
\end{equation}
 We update this estimate online using an auto-regressive approach. The offline calibration process is illustrated in Fig.~\ref{fig:wifi_to_farecard}.
\begin{figure}[!htb]
	\centering
	\includegraphics[width=0.6\columnwidth]{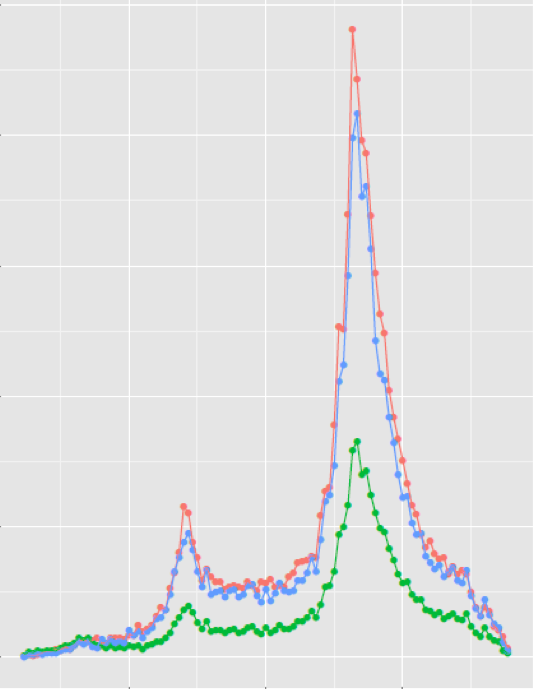}
	\caption{Number of passengers entering the station, 10 min discretization over a day (x-axis):  ticketing data considered as ground-truth (red), number of distinct connected device observations (green), and estimated number of passengers entering the station from connected devices (blue) using a linear scaling factor.}
	\label{fig:wifi_to_farecard}
\end{figure}   

Using the learned scaling factor, the count of passengers waiting to board a train can be derived from the number of observations from connected devices at any point in time, and in particular before and after a train arrives. More complex online or noise-modeling methods could  be considered  to improve the accuracy of the scaling factor.

Another relevant feature is the commuter wait time, which can be directly estimated by considering robust statistics of the time difference between the last observed record and the first observed record from connected devices who are observed to be traveling. In the next section we explain how we use these features in order to estimate the demand-supply gap.
\end{subsection}
\begin{subsection}{Demand-supply gap (DSG) estimation}
We propose using a discriminative classification method in order to estimate the demand-supply gap based on observations from connected devices. We consider the following feature set:
\begin{itemize}
\item count of commuters waiting to board a train,
\item count of commuters ``missing the train'', i.e. observed continuing to wait for the next train once a train departs,
\item waiting time third quartile,
\item waiting time standard deviation,
\item train headway obtained from the robust spectral clustering method.
\end{itemize}
We highlight that we are trying to estimate the macroscopic demand-supply gap, and not whether specific commuters will be left behind. We use greedy forward feature selection to select the most relevant features for model building. Since the datasets are highly skewed, with the vast majority of samples reflecting no DSG occurrences, we invoke a bootstrapping procedure to obtain an unbiased classification result.

Given the low number of DSG events, it is unrealistic to rely solely on station-specific models for accurate estimation. On the other hand, given the lack of stationarity of the underlying processes across stations and times, we cannot readily train models across the entire dataset. We normalize the features across the entire dataset, and then build a hierarchical model.

We train the models in a top-down fashion starting from a network-wide model for all stations on the network to line specific models, with different models for each unique line on the network, and finally to fine-grained models for each unique station on the network.
\end{subsection}

In the following section we present numerical results for both the movement model and the demand-supply gap model.
\end{section}
\begin{section}{Numerical results}\label{sec:mrt:evaluation}
\begin{subsection}{Train movement estimation}
\begin{subsubsection}{Performance metric}
To evaluate the performance of the train arrival detection method, we compare the estimated train arrivals derived by the baseline and the spectral clustering method to ground-truth arrival times at six stations. We use the following accuracy metrics:
\begin{itemize}
    \item hit rate: the proportion of arrival times within $1$ minute of the ground-truth, 
    \item root mean squared error (RMSE) of arrival time, expressed in minutes.
\end{itemize}
We first evaluate the methods under typical conditions, and then during an incident, during which train movements differ from their regular schedule.
\end{subsubsection}
\begin{subsubsection}{Typical conditions}
We analyze the results during typical conditions during the most challenging peak travel times, with short train headways, over the five workdays of a given week. Table~\ref{tab:mrt:comp-normal} presents the hit rate and RMSE.
\begin{table}[!htb]
    \centering
    \begin{tabular}{ccccc}
        \toprule
        Station ID&\multicolumn{2}{c}{Hit rate}&\multicolumn{2}{c}{RMSE}\\
        \cline{2-5}
        & Baseline & SC & Baseline & SC \\
        \midrule
        A  & 0.80 & \textbf{0.89} & 0.75 & \textbf{0.51} \\
        B  & \textbf{0.92} & 0.91 & \textbf{0.23} & 0.26 \\
        C & 0.96 & 0.96 & 0.21 & 0.21 \\
        D & 0.85 & \textbf{0.88} & 0.30 & \textbf{0.26} \\
        E & 1.00 & 1.00 & 0.00 & 0.00 \\
        F & 1.00 & 1.00 & 0.00 & 0.00 \\
        \bottomrule
    \end{tabular}
        \caption{Hit rate and RMSE: metrics are averaged over the peak hours of five workdays.}
    \label{tab:mrt:comp-normal}
\end{table}

Best performance (excluding ties) is marked in bold. Both methods perform well under these circumstances, though  spectral clustering moderately outperforms the baseline method.
\end{subsubsection}
\begin{subsubsection}{Incident scenario}
We further evaluate the two methods during a two-hour train incident. Specifically, we consider a real incident during which a train disruption occurs from around 7:15 to 8:15, illustrated in Figure~\ref{fig:mrt:incident}. The traffic is interrupted at station $15$ for half an hour and then partially resumes but remains perturbed until $9$am.
\begin{figure}[!htb]
 \centering
 \includegraphics[width =0.9\columnwidth]{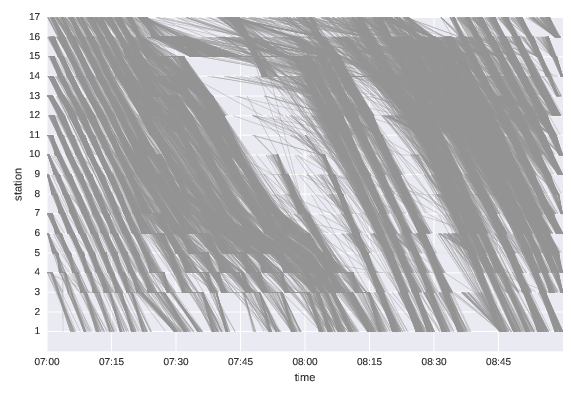}
 \caption{Commuter traces during a train incident.}
 \label{fig:mrt:incident}
\end{figure}

Figure~\ref{fig:mrt:incident} depicts commuter traces during an incident. The data being extremely noisy, journeys can easily be mis-identified, and clusters mis-detected. We analyze the performance of both the spectral clustering model and the baseline model over this time period.

Table~\ref{tab:mrt:comp-incident} shows the hit rate and RMSE at Station $15$ (chosen because ground truth at other stations is unavailable).
\begin{table}[!htb]
 \centering
 \begin{tabular}{cccc}
 \toprule
 \multicolumn{2}{c}{Hit 
 rate}&\multicolumn{2}{c}{RMSE}\\
 \midrule
 Baseline&Spectral clustering&Baseline&Spectral clustering\\
 \midrule
 0.06& \textbf{ 0.52 } & 7.50 & \textbf{ 3.63 } \\
 \bottomrule
 \end{tabular}
  \caption{Hit rate and RMSE during a train incident:  metrics are averaged from $7$ to $9$ am at station $15$.}
 \label{tab:mrt:comp-incident}
\end{table}
In this case of complex movements with overlapping clusters, the spectral clustering method significantly outperforms the baseline approach, by a factor $10$ in terms of hit rate, and by a factor $2$ for the RMSE metric. 
\end{subsubsection} 
\end{subsection}
\begin{subsection}{Demand-supply gap model evaluation}
\begin{subsubsection}{Performance metric}
We consider a performance metric based on the DSG value over $30$ minute time intervals, as this is how it is measured by transport operators. With the objective of making results easier to interpret, we express the DSG as the percentage of commuters left behind, i.e. unable to board the first train they waited for, over the set of commuters who intended to board during the $30$ minute interval.

We first present the model parameters obtained, then results of the detection of occurrences of DSG, and finally we provide more details on the accuracy of the DSG estimates and its robustness properties.
\end{subsubsection}
\begin{subsubsection}{Model training}
We make use of  100 K  ground-truth DSG event labels (positive and negative instances) collected over a period of $8$ months at  $60$ stations. The dataset is highly skewed with a majority of non-DSG occurrences. The associated raw wifi traces dataset consists of  $1$ GB of raw wifi traces per day, or about $10$ billion records over the period considered. The dataset is split into $75\%$ training and $25\%$ testing. 

We perform $10$-fold cross validation. Outlier Winsorization is performed with a percentile of $0.99$. We build a GLM model with logistic regression in R. The F1 score is used as the metric in training.  A grid search reveals a probability cutoff threshold of $0.1$ on the cross validation set. The test set  yields a precision of $84\%$ and a recall of $75\%$. Other models examined including SVM, Random forests and LogitBoost did not improve  accuracy. 

In Table~\ref{table:utbModelWeights} we present the feature weights obtained for a logistic regression model focused on detecting the severity of the DSG measured as the average number of trains an impacted passenger  had to miss. 
  \begin{table}[!htb]
	\centering
	\begin{tabular}{cc}
		\toprule
		Feature & Weight\\
		\midrule
		Intercept & -7.3645 \\
		Count & 0.00033 \\
		Missed Count & 0.06824 \\
		75\% Quantile WaitTime & 0.00682 \\
		Std Dev WaitTime  & 0.00014 \\
		Headway time & 0.00251\\
		\bottomrule
	\end{tabular}
	\caption{Feature weights: weights from logistic regression model to classify an occurrence of a train arrival as DSG or not.}
	\label{table:utbModelWeights}
\end{table}

It is  clear  that the results are most sensitive to the feature indicating the number of commuters still waiting to board after the train departs, as this feature is a noisy aggregate observation of the DSG. We observe that other features such as wait time statistics have a reduced albeit significant impact. This is explained by the inherent noise in the count estimates while time-based estimates are less subject to non-stationary sampling rate, hence more stable by construction.
\end{subsubsection}
\begin{subsubsection}{Model testing}
We now analyze the benefit of the hierarchical model structure. In Table~\ref{table:modelHierarchy} we present the Precision, Recall, and Accuracy of detecting DSG with specific models.
\begin{table}[!htb]
	\centering
	\begin{tabular}{ccccc}
		\toprule
		Category & $\#$Models & Precision & Recall & Accuracy\\
		\midrule
		Network   & 1 & 75 & 72 & 98\\
		Line    & 1 to 10 & 77 & 72 & 98\\
		Station & 10 to 100 & 85 & 75 & 99\\
		\bottomrule
	\end{tabular}
	\caption{Model hierarchy: complexity and performance of different model categories.}
	\label{table:modelHierarchy}
\end{table}

 Observe that the use of  specific models moderately improves  model accuracy, in particular in terms of reducing the number of false alarms. On the other hand, it significantly increases (by 2 orders of magnitude) the complexity associated with model maintenance, i.e. automated training, deployment, and monitoring. As these algorithms form the basis of a large-scale, real-time system, the importance of model maintenance is not to be neglected.

Overall accuracy results are presented in Figure~\ref{fig:UTB1_mae} and Figure~\ref{fig:UTB2plus_mae} in terms of the mean absolute error of DSG values. We further split the DSG  into values whereby left behind passengers only miss $1$ train (DSG $1$), or miss $2$ or more trains (DSG $2+$). The error statistics are reported at a set of $7$ stations where a DSG is observed.
\begin{figure}[!htb]
	\centering
	\includegraphics[width =0.7\columnwidth]{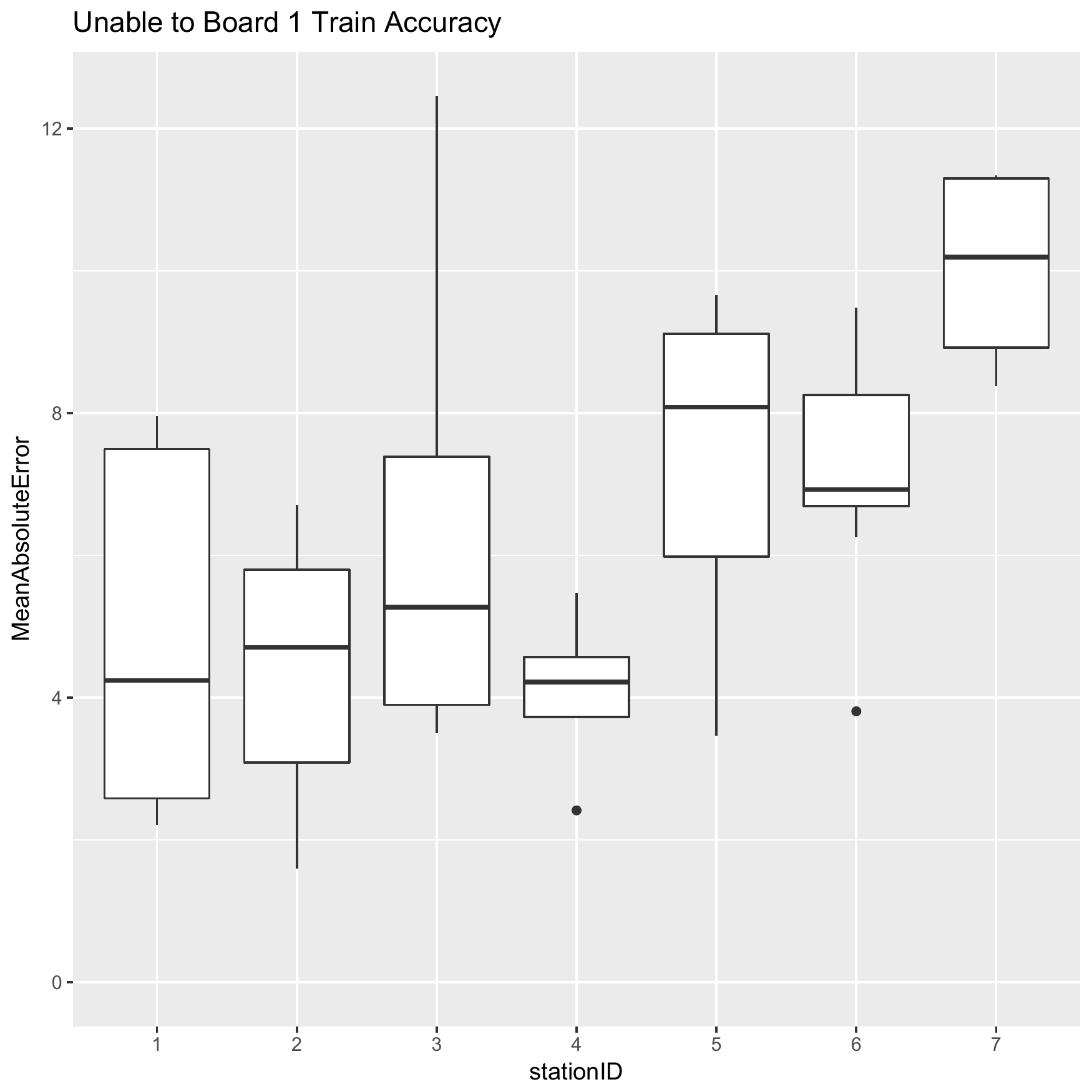}
	\caption{Accuracy of the demand-supply gap model for DSG of 1 train.}
	\label{fig:UTB1_mae}
\end{figure}
\begin{figure}[!htb]
	\centering
	\includegraphics[width = 0.7\columnwidth]{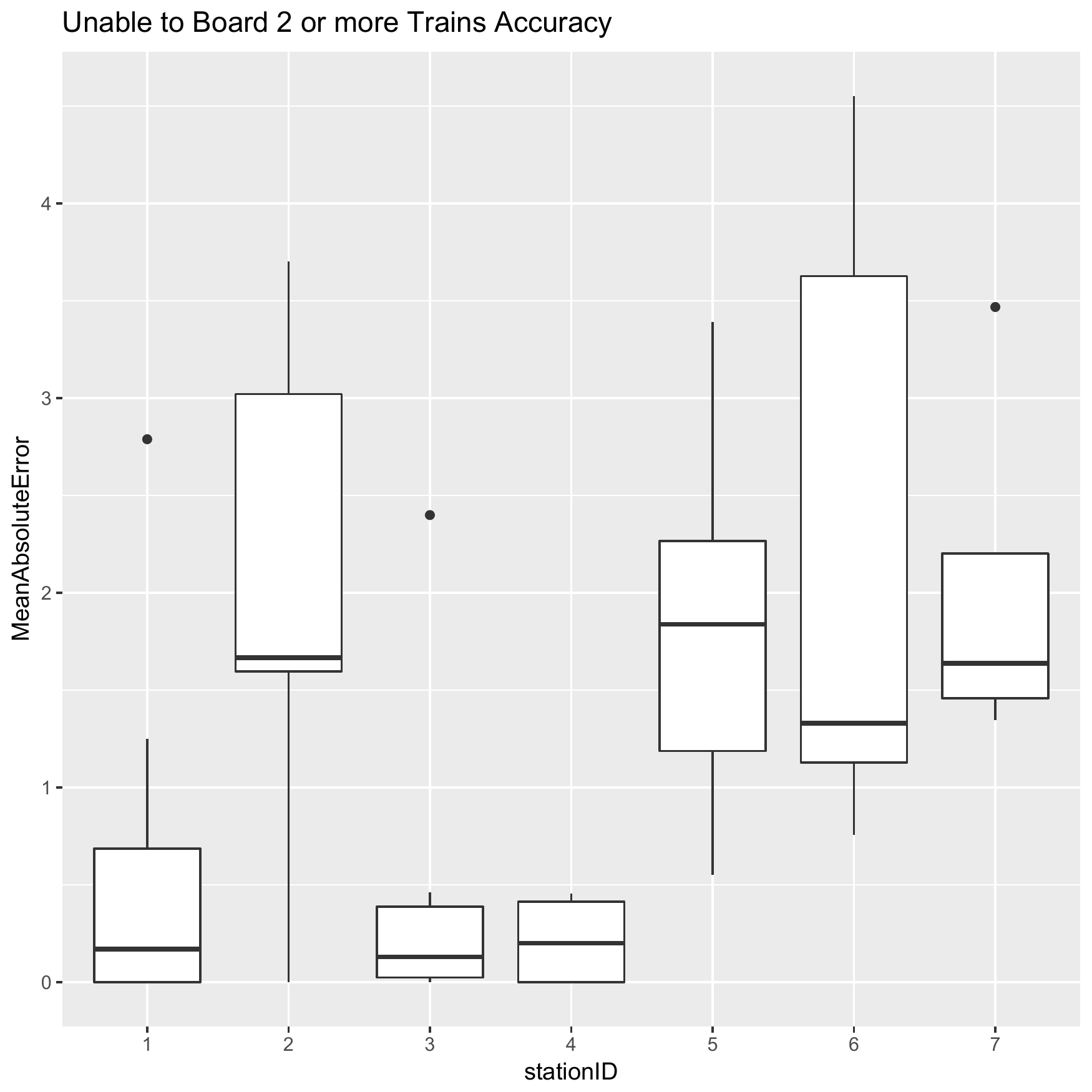}
	\caption{Accuracy of the demand-supply gap model for DSG of 2 or more trains.}
	\label{fig:UTB2plus_mae}
\end{figure}

The inherent variability due to non-stationarity of the overall dataset is visible from the results. The median error remains relatively low for most stations.
\end{subsubsection}
\begin{subsubsection}{Model robustness}
Given the inherent noise of the data, in particular as regards the variable proportion of commuters  observed, we    further analyze the robustness of the model estimates to that proportion:
\begin{itemize}
\item we train the model on the full dataset, i.e.  a sampling factor of $100\%$. Recall that the true sampling factor, with respect to the number of commuters, is unknown, time-varying and location-dependent.
\item we test the model with subsets of the data obtained by down-sampling the full dataset. We down-sample by increments of $10\%$ over the interval $[0,100]$.
\end{itemize}
Figure~\ref{fig:robustness_sampling_levels} presents the average Precision and Recall as a function of the down-sampling value. 
\begin{figure}[!htb]
	\centering
	\includegraphics[width=0.8\columnwidth]{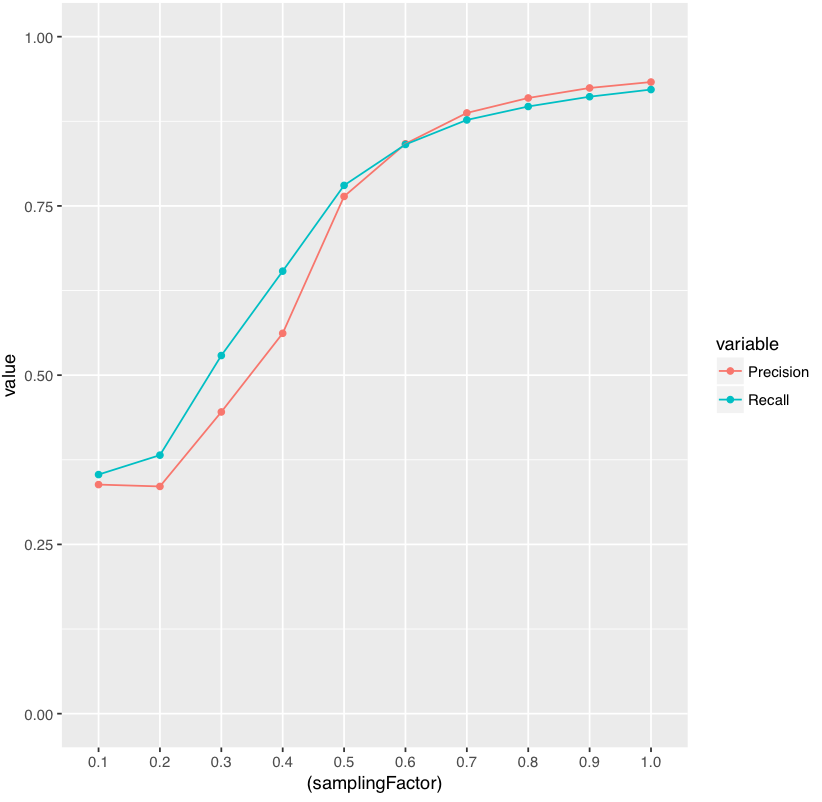}
	\caption{Precision and recall for the DSG model under different sampling factors.}
	\label{fig:robustness_sampling_levels}
\end{figure} 
We observe that the model is able to maintain good performance when the sampling factor is above $60\%$.
\end{subsubsection}
\end{subsection}
\end{section}
\begin{section}{Conclusion}\label{sec:conc}
In this work we considered the problem of large-scale inference of public transport level of service based on connected devices data. We investigated the specific properties of such datasets, shared in particular with IoT data. We highlighted the inherent noise of the data, not only in terms of spatio-temporal accuracy of the data, but also in terms of the non-stationarity of the underlying sensing scheme such as its sampling factor.

We introduced unsupervised learning methods appropriate for estimating global conditions of the public transport network such as train headways and train line level of service. We further developed robust classification methods using estimated train movements as a building block in order to estimate the demand-supply gap in real-time and in a tractable manner. We emphasize that the demand-supply gap is a very general quantity of relevance outside of public transport.

Numerous extensions can be considered. First it is clear that more complex approaches can be developed for the training phase. In order to compensate the impact of data uncertainty, one could for instance jointly consider all network events as informative of the conditions at any specific station. Second, it is clear that a key aspect of the deployment of such systems is the ability to monitor  model stability in real-time and re-train if needed. Lastly, in order to support more diverse crowd monitoring applications, it would be of interest to develop similar  inference processes  in two and  three-dimensional space.
\end{section}
\bibliographystyle{plain}
\bibliography{bib_roma}
\end{document}